\documentclass[letterpaper]{article} % DO NOT CHANGE THIS
\usepackage[]{aaai23}
\usepackage{times}  % DO NOT CHANGE THIS
\usepackage{helvet}  % DO NOT CHANGE THIS
\usepackage{courier}  % DO NOT CHANGE THIS
\usepackage[hyphens]{url}  % DO NOT CHANGE THIS
\usepackage{graphicx} % DO NOT CHANGE THIS
\usepackage{natbib}  % DO NOT CHANGE THIS AND DO NOT ADD ANY OPTIONS TO IT
\usepackage{caption} % DO NOT CHANGE THIS AND DO NOT ADD ANY OPTIONS TO IT
\usepackage{algorithm}
\usepackage{algorithmic}

\usepackage{newfloat}
\usepackage{listings}
\DeclareCaptionStyle{ruled}{labelfont=normalfont,labelsep=colon,strut=off} % DO NOT CHANGE THIS
\floatstyle{ruled}
\newfloat{listing}{tb}{lst}{}
\floatname{listing}{Listing}
\usepackage{amsfonts}

\title{Reducing Over-smoothing in Graph Neural Networks Using Relational Embeddings}
\author{
    Yeskendir Koishekenov
}
\affiliations{
    University of Amsterdam \\
    yeskendir.koishekenov@student.uva.nl
}

\begin{document}

\maketitle

\begin{abstract}
Graph Neural Networks (GNNs) have achieved a lot of success with graph-structured data. However, it is observed that the performance of GNNs does not improve (or even worsen) as the number of layers increases. This effect has known as over-smoothing, which means that the representations of the graph nodes of different classes would become indistinguishable when stacking multiple layers. In this work, we propose a new simple, and efficient method to alleviate the effect of the over-smoothing problem in GNNs by explicitly using relations between node embeddings. Experiments on real-world datasets demonstrate that utilizing node embedding relations makes GNN models such as Graph Attention Network more robust to over-smoothing and achieves better performance with deeper GNNs. Our method can be used in combination with other methods to give the best performance. GNN applications are endless and depend on the user's objective and the type of data that they possess. Solving over-smoothing issues can potentially improve the performance of models on all these tasks.
\end{abstract}

\section{Introduction}

Graph neural networks (GNNs) are a family of neural networks that can learn from graph-structured data. Starting with the success of GCN \cite{kipf2016semi} in achieving state-of-the-art performance on semi-supervised classification, several variants of GNNs have been developed for this task, including Graph-SAGE \cite{hamilton2017inductive}, GAT \cite{velivckovic2017graph}, GATv2 \cite{brody2021attentive}, EGNN \cite{satorras2021n} to name a few most recent ones.

A key issue with GNNs is their depth limitations. It has been observed that stacking the layers often results in significantly worse performance for GNNs, such as GCN and GAT. One of the factors associated with this performance drop is the phenomenon called \textit{over-smoothing}. The first to call attention to the over-smoothing problem was in the work of \citet{li2018deeper}. Having shown that the graph convolution is a type of Laplacian smoothing, they proved that after repeatedly applying Laplacian smoothing many times, the features of the nodes in the (connected) graph would converge to similar values. Later, several other works have alluded to the same problem \cite{li2019deepgcns, luan2019break}.

The main research question of this paper is how to keep two node representations distinguishable as we increase their receptive fields. We propose that stressing the difference between two nodes will prevent their embeddings from becoming too similar. The main contributions of this paper are:
\begin{itemize}
    % 1: proposed new method 
    \item A method to reduce the over-smoothing in GNNs, specifically in GAT, by using not only node embeddings as features but also explicitly using their relations. To validate our method, we use different metrics that quantify the over-smoothing phenomenon.
    % 2: new method alleviates over-smoothing
    \item We empirically show that deeper GAT equipped with our proposed method improves node classification accuracy in a real-world scenario where graphs have missing node features.
    \item Improve other approaches to tackling over-smoothing by employing our method.
\end{itemize}

\section{Related Work}
\subsection{Over-smoothing}
A straightforward way to reduce the effect of over-smoothing is to simply reduce the number of layers. However, this implies not exploiting the multi-hop information in the case of complex-structured data and consequently limiting end-task performance. Therefore, having over-smoothing as an issue, researchers encounter a trade-off between a low-efficiency model and a model with more depth but less expressivity in terms of node representations. \citet{oono2019graph}, \citet{cai2020note} did extensive analysis of expressive power in GNN.

There have been several attempts to make GNNs more robust to over-smoothing. \citet{xu2018representation} introduced Jumping Knowledge Networks, which employ skip connections for multi-hop message passing and also enable different neighborhood ranges. \citet{klicpera2019combining} proposed a propagation scheme based on personalized PageRank that ensures locality (via teleports) which in turn prevents over-smoothing. \citet{li2019deepgcns} built on ideas from ResNet to use residual as well as dense connections to train deep GCNs. \citet{rong2019dropedge} proposed DropEdge to alleviate over-smoothing through message passing reduction via removing a certain fraction of edges at random from the input graph. 
% \cite{zhao2019pairnorm} and \cite{zhou2020towards} proposed normalization layers to prevent node embeddings from becoming too similar. 

Recently, utilizing normalization layers showed effectiveness in preventing node embeddings from becoming too similar. \citet{zhao2019pairnorm} proposed PairNorm, a normalization scheme that ensures that the total pairwise node feature distances remain constant across layers. \citet{zhou2020towards} introduced DGN which clusters nodes and prevents distinct groups from having close features.

It is essential to quantify over-smoothing to validate solutions. The main goal is to improve or at least prevent the drop in accuracy as the number of layers increases. However, different factors besides over-smoothing can impact it. Therefore, proposed approaches should be validated on various quantitative metrics that directly measure over-smoothing in graphs such as group distance ratio and Instance information gain \cite{zhou2020towards}, or row-diff. and col-diff. \cite{zhao2019pairnorm}.

\subsection{Node Relations}
In Natural Language Processing tasks such as Natural Language Inference, understanding entailment, and contradiction, sentence embedding relations were used to distinguish two vector representations \cite{conneau2017supervised, nie2019dissent}. For example, \citet{conneau2017supervised} applied 3 matching methods to extract relations between two sentence embeddings: concatenation of the two representations, element-wise product, and absolute element-wise difference. In a similar manner \citet{nie2019dissent} in addition to sentence embeddings used a fixed set of common pair-wise vector operations: subtraction, multiplication, and average. They showed empirically that non-linear interactions between feature vectors are needed. Motivated by these works, we apply similar techniques in GNNs to disentangle node representations hence alleviating over-smoothing.

\section{Preliminaries}

In this work, we consider the semi-supervised node classification task as an example and illustrate how to handle the over-smoothing issue. A graph is represented by $G={\mathcal{V}, \mathcal{E}}$, where $\mathcal{V}$ and $\mathcal{E}$ represent the sets of nodes and edges, respectively. Each node $i \in \mathcal{V}$ is associated with a feature vector $x_{i} \in \mathbb{R}^{d}$  and $X = [x_{1}, ..., x_{n}]^{T}$ denotes the feature matrix, and a subset $\mathcal{V}_{l} \subset \mathcal{V}$ of the nodes are labeled, i.e. $y_{i} \in 1, ..., C$ for each $i \in \mathcal{V}$ where $C$ is the number of classes. The task is to learn a hypothesis that predicts $y_{i}$ from $x_{i}$ that generalizes to the unlabeled nodes $\mathcal{V}_{u} = \mathcal{V}\setminus\mathcal{V}_{l}$.

\subsection{Graph Neural Network}

A GNN layer updates every node representation by aggregating its neighbors' representations. A layer's input is a set of node representations $\{h_{i} \in \mathbb{R}^{d} | i \in \mathcal{V}\}$ and the set of edges $\mathcal{E}$. A layer outputs a new set of node representations $\{h^{'}_{i} \in \mathbb{R}^{d} | i \in \mathcal{V}\}$, where the same parametric function is applied to every node given its neighbors $\mathcal{N}_{i} = \{j \in \mathcal{E} | (j,i) \in \mathcal{E} \}$:
\begin{equation}
\label{eqn:gnn_eqn}
h^{'}_{i} = f_{\theta}(h_{i}, Aggregate({h_{j} | j \in \mathcal{N}_{i}}))
\end{equation}
The design of $f$ and $Aggregate$ is what mostly distinguishes one type of GNN from the other. For example, GAT \cite{velivckovic2017graph} instantiates Equation \ref{eqn:gnn_eqn} by computing a learned weighted average of the representations of $\mathcal{N}_{i}$. A scoring function $e:\mathbb{R}^{d} \times \mathbb{R}^{d} \rightarrow \mathbb{R}$ computes a score for every edge $(j, i)$, which indicates the importance of the features of the neighbor $j$ to the node $i$:
\begin{equation}
\label{eqn:e_eqn}
e_{i, j} = \sigma(a^{T} \cdot [Wh_{i}||Wh_{j}])
\end{equation}
where $a\in\mathbb{R}^{2d'}, W\in\mathbb{R}^{d'\times d}$ are parameters learned, $\sigma$ is a non-linear activation function (e.g. $LeakyReLU$), and $||$ denotes vector concatenation. These attention scores are normalized across all neighbors $j\in\mathcal{N}_{i}$ using softmax, and the attention function is defined as:
\begin{equation}
\label{eqn:alpha_eqn}
\alpha_{ij} = softmax_{j}(e_{i,j}) = \frac{exp(e_{i,j})}{\sum_{j'\in\mathcal{N}_{i}}exp(e_{i, j'})}
\end{equation}
Then GAT computes a weighted average of the transformed features of the neighbor nodes (followed by nonlinearity $\sigma$) as the new representation of $i$, using the normalized attention coefficients:

\begin{equation}
\label{eqn:h_gat_eqn}
h_{i}^{'} = \sigma(\sum_{j\in\mathcal{N}_{i}} \alpha_{ij} \cdot Wh_{j}).
\end{equation}

\section{Approach}

In this work, we propose to alleviate the over-smoothing problem by utilizing node relations. In the message-passing framework, each node aggregates feature vectors from neighboring nodes via a permutation equivariant function and outputs a message vector. Then it updates its own embedding using this message vector. Traditionally, the message vector is computed as in Equations \ref{eqn:e_eqn} - \ref{eqn:h_gat_eqn}. We propose to additionally utilize the relation of two node embeddings. In this work, we will empirically search for the best pair-wise vector operation between two embeddings that can help to solve the problem. We extend Equation \ref{eqn:e_eqn} to utilize the relation of node embeddings as shown below:

\begin{equation}
\label{eqn:new_e_eqn}
e_{i, j} = \sigma(a^{T} \, [W^{'}h_{i}||W^{'}h_{j}||W^{''}\, relation(h_{i}, h_{j})])
\end{equation}

where $W^{'}$ and $W^{''}$ parameterize embeddings and their relations. The intuitive explanation of our approach is that utilizing node relation information stresses the difference between two nodes, which in return will prevent node embeddings from becoming indistinguishable. Analogous approaches were used in Natural Language Inference task \cite{conneau2017supervised, nie2019dissent} to distinguish two vector representations. 

We experiment with four matching methods, $relation(h_{i}, h_{j})$ in Equation \ref{eqn:new_e_eqn}, to extract relations between two nodes: 
% (a) difference $h_{i}-h_{j}$; (b) absolute difference $|h_{i}-h_{j}|$; (c) element-wise difference $h_{i}*h_{j}$; and (d) concatenation of absolute difference and element-wise product.
\begin{itemize}
    \item difference: $h_{i}-h_{j}$
    \item absolute difference: $|h_{i}-h_{j}|$
    \item element-wise product: $h_{i}*h_{j}$
    \item concatenation of absolute difference and element-wise product
\end{itemize}

\begin{figure}
    \centering
    \includegraphics[width=0.9\columnwidth]{"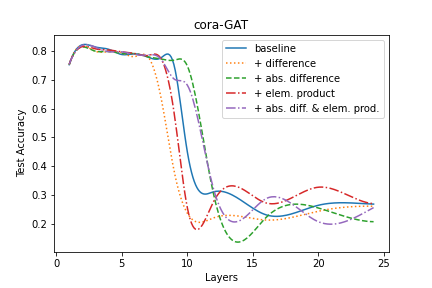"}
    \caption{GAT's test accuracy with an increasing number of layers on the Cora dataset.}
    \label{fig:test_all}
\end{figure}

\begin{figure}
    \centering
    \includegraphics[width=0.8\columnwidth]{"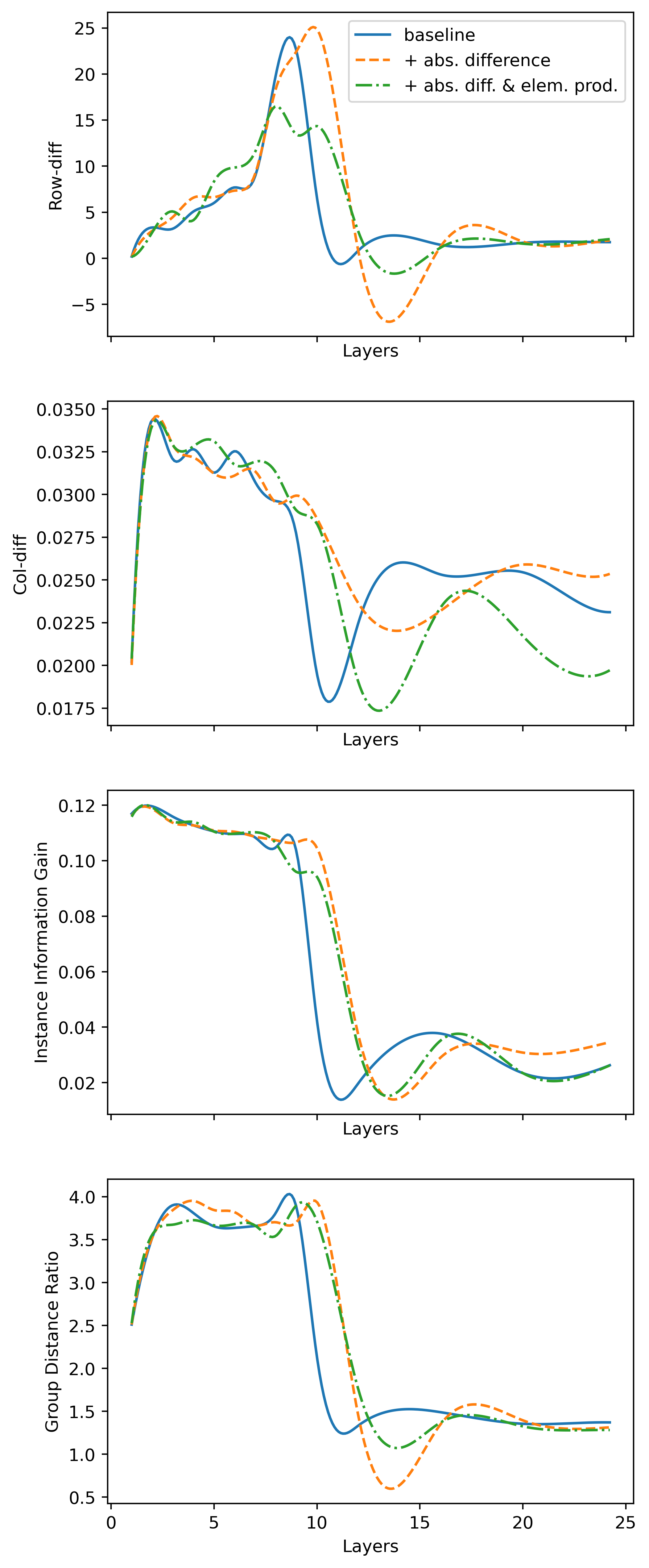"}
    \caption{The row-diff., col-diff., instance information gain, and group distance ratio of GAT on Cora dataset with an increasing number of layers}
    \label{fig:cora_GAT_os}
\end{figure}

\section{Experiments}

We now empirically evaluate the effectiveness of our method on real-world datasets.

\subsection{Experiment Setup}

\paragraph{Datasets} 
We use a well-known benchmark dataset in GNN domain:  \textit{Cora}, \textit{Pubmed}, and \textit{Citeseer} \cite{yang2016revisiting}. These citation network datasets contain sparse bag-of-words feature vectors for each document (node) and a list of citation links (edges) between documents. We treat the citation links as (undirected) edges and construct a binary, symmetric adjacency matrix. Each document has a class label. We use the same data splits as \citet{kipf2016semi}, where all nodes outside the train and validation are used as a test set. 

\paragraph{Implementations} 
As a base model, we use GAT \cite{velivckovic2017graph}. Following the previous settings, we choose the hyperparameters of the model and optimizer as follows. We set the number of hidden units to 64 and the number of attention heads in GAT is 1. During training, we use the Adam optimizer \cite{kingma2014adam} with a dropout rate of 0.6, weight decay 5e-4 ($L_{2}$ regularization). In experiments with PairNorm \cite{zhao2019pairnorm} and DGN \cite{zhou2020towards}, we use exactly the same hyperparameters provided in authors' papers or released code. We run each experiment on NVIDIA TITAN RTX within 1000 epochs 4 times with different seeds and report the average performance.

% \begin{figure}
%     \centering
%     \includegraphics[width=0.9\columnwidth]{"plots/aaai\_test\_acc\_all.png"}
%     \caption{GAT's test accuracy with an increasing number of layers on Cora dataset.}
%     \label{fig:test_all}
% \end{figure}

\begin{table*}[htp]
\centering
\resizebox{2\columnwidth}{!}{
\begin{tabular}{l|l|l l|l l|l l|l l|l l|l l}
     & Missing Percentage & \multicolumn{2}{c|}{0} & \multicolumn{2}{c|}{20} & \multicolumn{2}{c|}{40} & \multicolumn{2}{c|}{60} & \multicolumn{2}{c|}{80} & \multicolumn{2}{c}{100} \\
    Dataset & Relational Embedding & Acc.& \#L & Acc.& \#L & Acc. & \#L & 
    Acc. & \#L & Acc. & \#L & Acc. & \#L \\
    \hline
    \hline
    Cora & none & 81.903 & 3 & \textbf{81.262} & 2 & 79,05 & 2 & 77,539 & 3 & 74,758 & 4 & 71,639 & 8 \\
    * & + abs. difference & \textbf{82.374} & 2 & 81.25 & 2 & 79.171 & 2 & 77.563 & 4 & 75.0 & 7 & 71.434 & 9 \\
    * & + abs. diff \& elem. prod. & 81.854 & 3 & 80.936 & 2 & \textbf{79.678} & 4 & \textbf{77.817} & 4 & \textbf{75.459} & 4 & \textbf{71.954} & 8 \\
    \hline
    Pubmed & none & \textbf{77.34} & 2 & \textbf{77.611} & 2 & 77.5 & 7 & 77.095 & 7 & 76.784 & 7 & 70.187 & 8 \\
    * & + abs. difference & 77.32 & 8 & 77.123 & 2 & 77.465 & 7 & 77.708 & 8 & 77.1 & 8 & 69.822 & 7 \\
    * & + abs. diff \& elem. prod. & 77.042 & 6 & 77.327 & 8 & \textbf{77.519} & 8 & \textbf{77.691} & 8 & \textbf{77.664} & 10 & \textbf{71.334} & 10 \\
    \hline
    Citeseer & none & \textbf{68.563} & 2 & 67.187 & 2 & 63.881 & 2 & 60.251 & 4 & 55.006 & 5 & \textbf{46.417} & 6 \\
    * & + abs. difference & 69.025 & 2 & 67.603 & 2 & 63.613 & 3 & \textbf{61.119} & 4 & 55.218 & 4 & 46.204 & 6 \\
    * & + abs. diff \& elem. prod. & 69.256 & 2 & \textbf{67.409} & 2 & \textbf{64.01} & 2 & 60.537 & 3 & \textbf{55.43} & 5 & 45.586 & 6 \\
    \hline
    
\end{tabular}
}
\caption{Comparison of test accuracies of GAT with and without relational embeddings with varying missing percentages on different datasets. \#L denotes the optimal layer numbers where the model achieves the highest performance.}
\label{cora_missig_rate}
\end{table*}

% tables here:

\begin{table*}[ht]
\centering
\resizebox{2\columnwidth}{!}{
\begin{tabular}{l|l|l l|l l|l l|l l|l l|l l}
     & Dataset &  \multicolumn{4}{c|}{Cora} & \multicolumn{4}{c|}{Pubmed} & \multicolumn{4}{c}{Citeseer} \\
    * & Missing Percentage & \multicolumn{2}{c|}{0} & \multicolumn{2}{c|}{100} & \multicolumn{2}{c|}{0} & \multicolumn{2}{c|}{100} & \multicolumn{2}{c|}{0} & \multicolumn{2}{c}{100} \\
    Method & Relational Embedding & Acc.& \#L & Acc.& \#L & Acc. & \#L & 
    Acc. & \#L & Acc. & \#L & Acc. & \#L \\
    \hline
    \hline
    PairNorm & none & \textbf{78.506} & 2 & 69.5 & 7 & \textbf{77.051} & 8 & 73.491 & 16 & 67.15 & 3 & 50.933 & 5 \\
    * & + abs. difference    & 78.433 & 2 & 70.43 & 6 & 76.707 & 8 & 73.345 & 16 & 67.205 & 1 & \textbf{51.487} & 5 \\
    * & + abs. diff \& elem. product & 78.179 & 2 & \textbf{70.636} & 4 & 76.848 & 10 & \textbf{73.833} & 16 & \textbf{67.732} & 1 & 51.321 & 5 \\
    \hline
    DGN & none & 81.093 & 2 & 69.85 & 4 & \textbf{77.259} & 4 & 63.481 & 4 & \textbf{67.723} & 2 & 48.735 & 4 \\
    * & + abs. difference  & 81.081 & 2 & 70.261 & 4 & 76.707 & 8 & \textbf{63.61} & 4 & 67.372 & 1 & \textbf{49.926} & 4 \\
    * & + abs. diff \& elem. product & \textbf{81.564} & 2 & \textbf{70.563} & 4 & 77.105 & 4 & 61.095 & 4 & 67.464 & 2 & 49.621 & 4 \\
    
\end{tabular}
}
\caption{Comparison of test accuracies of GAT with PairNorm and DGN normalizations with and without relational embeddings on different datasets. \#L denotes the optimal layer numbers where the model achieves the highest performance.}
\label{other_methods}
\end{table*}

\subsection{Measuring Over-smoothing}
In addition to standard test accuracy, we use the following metrics to quantify over-smoothing and validate our proposed approach. These metrics consider both pairwise and group information in graphs.

\paragraph{Row-diff and Col-diff}
\citet{zhao2019pairnorm} introduced two metrics to quantify node-wise and feature-wise over-smoothing. The row-diff measure is the average of all pairwise distances between the node features and quantifies node-wise over-smoothing. The col-diff is the average of all pairwise distances between the columns of the representation matrix and quantifies feature over-smooth

\paragraph{Group Distance Ratio and Instance Information Gain}
\citet{zhou2020towards} introduced two metrics: Group Distance Ratio, $R_{Group}$, and Instance Information Gain, $G_{Ins}$. Group Distance Ratio first clusters nodes of the same class label into a group to formulate the labeled node community, then measures the ratio of inter-group distance over intra-group distance in the Euclidean space. A small $R_{Group}$ leads to the over-smoothing issue where all groups are mixed together. Instance information gain, $G_{Ins}$, is defined as how much input feature information is contained in the final representation. $G_{Ins}$ measures the dependency between node feature and representation via their mutual information.

\subsection{Experiment Analysis}

\paragraph{Choosing effective relational embeddings}
We first show in Figure \ref{fig:test_all} the test accuracy of GAT on the \textit{Cora} dataset as we increase the number of layers with different relational embeddings. Employing node embedding relations such as their absolute difference and their concatenation with their element-wise product improves the performance of deeper models. We can notice that the baseline graph line was shifted in the right direction. Therefore, we do further analysis focusing on these node relation features.

\paragraph{Reducing over-smoothing} 
We show in Figure \ref{fig:cora_GAT_os} the metrics quantifying over-smoothing, i.e. row-diff, col-diff, instance information gain, group distance ratio, of GAT model on the \textit{Cora} dataset as we increase the number of layers with different relational embeddings. Here we observe the same trend with test accuracy that the baseline graph line for all metrics was shifted in the right direction by adding relational embeddings. In other words, the receptive field of the nodes increased while maintaining performance. It indicates that utilizing node relation features such as absolute difference or element-wise product of two embeddings shows improvement over the baseline. The improvement over four different metrics quantifying over-smoothing supports that our method reduces the representation similarity both between groups and pairs of nodes hence alleviating the over-smoothness of nodes over a graph.

\paragraph{Case where deeper is better}
We demonstrated that our method makes deeper models more robust to over-smoothing. However, the overall test accuracy did not improve significantly. This is due to the fact that architectures with no more than 2–4 layers are sufficient for the popular graph benchmark datasets. Our method shows its power in a setting where it required a large number of layers to achieve its best performance. One example is the real-world scenario when a notable portion of the nodes lack feature vectors. This variant of a task is called semi-supervised node classification with missing vectors \cite{zhao2019pairnorm}. In Table \ref{cora_missig_rate} we show the global best test accuracy of GAT on the Cora, Pubmed, and Citeseer, datasets along with the optimal layer number \#L under varying feature missing rates. As the percentage of missing rates increase the GAT model utilizing node relations consistently outperforms its vanilla version by tackling over-smoothing. As we mentioned, in setting with missing feature vectors deeper models achieve the best performance. 

\paragraph{Improving other methods with node relational embeddings}
Until now, we showed that the model robustness to over-smoothing increases when it uses relational embeddings. However, the absolute results are not better than alternative methods tackling over-smoothing such as PairNorm \cite{zhao2019pairnorm} or DGN \cite{zhou2020towards}. The strong advantage of our approach is that it can be easily used in the combination with other methods to bolster their performance. Table \ref{other_methods} shows the performance of other methods with and without relational embeddings. We can see that our method is most effective in the case of missing node features.

\section{Conclusioin}
In this work, we proposed a new simple but effective approach to reducing the impact of over-smoothing on training graph neural networks. We showed that utilizing some non-linear interactions between node embeddings such as absolute difference and element-wise product can mitigate the effect of the over-smoothing problem in the Graph Attention Network. To achieve it we validated our approach not only on accuracy, but also on "over-smoothing"-specific metrics such as row-diff., col-diff., instance information gain, and group distance ratio. Our approach also showed its effectiveness in combination with other methods to bolster their performance.

% Use \bibliography{yourbibfile} instead or the References section will not appear in your paper
% \nobibliography{aaai23}
% \bibliography{aaai23.bib}

\bibliography{aaai23.bib}

\begin{thebibliography}{19}
\providecommand{\natexlab}[1]{#1}

\bibitem[{Brody, Alon, and Yahav(2021)}]{brody2021attentive}
Brody, S.; Alon, U.; and Yahav, E. 2021.
\newblock How attentive are graph attention networks?
\newblock \emph{arXiv preprint arXiv:2105.14491}.

\bibitem[{Cai and Wang(2020)}]{cai2020note}
Cai, C.; and Wang, Y. 2020.
\newblock A note on over-smoothing for graph neural networks.
\newblock \emph{arXiv preprint arXiv:2006.13318}.

\bibitem[{Conneau et~al.(2017)Conneau, Kiela, Schwenk, Barrault, and
  Bordes}]{conneau2017supervised}
Conneau, A.; Kiela, D.; Schwenk, H.; Barrault, L.; and Bordes, A. 2017.
\newblock Supervised learning of universal sentence representations from
  natural language inference data.
\newblock \emph{arXiv preprint arXiv:1705.02364}.

\bibitem[{Hamilton, Ying, and Leskovec(2017)}]{hamilton2017inductive}
Hamilton, W.; Ying, Z.; and Leskovec, J. 2017.
\newblock Inductive representation learning on large graphs.
\newblock \emph{Advances in neural information processing systems}, 30.

\bibitem[{Kingma and Ba(2014)}]{kingma2014adam}
Kingma, D.~P.; and Ba, J. 2014.
\newblock Adam: A method for stochastic optimization.
\newblock \emph{arXiv preprint arXiv:1412.6980}.

\bibitem[{Kipf and Welling(2016)}]{kipf2016semi}
Kipf, T.~N.; and Welling, M. 2016.
\newblock Semi-supervised classification with graph convolutional networks.
\newblock \emph{arXiv preprint arXiv:1609.02907}.

\bibitem[{Klicpera, Bojchevski, and
  G{\"u}nnemann(2019)}]{klicpera2019combining}
Klicpera, J.; Bojchevski, A.; and G{\"u}nnemann, S. 2019.
\newblock Combining neural networks with personalized pagerank for
  classification on graphs.
\newblock In \emph{International conference on learning representations}.

\bibitem[{Li et~al.(2019)Li, Muller, Thabet, and Ghanem}]{li2019deepgcns}
Li, G.; Muller, M.; Thabet, A.; and Ghanem, B. 2019.
\newblock Deepgcns: Can gcns go as deep as cnns?
\newblock In \emph{Proceedings of the IEEE/CVF international conference on
  computer vision}, 9267--9276.

\bibitem[{Li, Han, and Wu(2018)}]{li2018deeper}
Li, Q.; Han, Z.; and Wu, X.-M. 2018.
\newblock Deeper insights into graph convolutional networks for semi-supervised
  learning.
\newblock In \emph{Thirty-Second AAAI conference on artificial intelligence}.

\bibitem[{Luan et~al.(2019)Luan, Zhao, Chang, and Precup}]{luan2019break}
Luan, S.; Zhao, M.; Chang, X.-W.; and Precup, D. 2019.
\newblock Break the ceiling: Stronger multi-scale deep graph convolutional
  networks.
\newblock \emph{Advances in neural information processing systems}, 32.

\bibitem[{Nie, Bennett, and Goodman(2019)}]{nie2019dissent}
Nie, A.; Bennett, E.; and Goodman, N. 2019.
\newblock DisSent: Learning sentence representations from explicit discourse
  relations.
\newblock In \emph{Proceedings of the 57th Annual Meeting of the Association
  for Computational Linguistics}, 4497--4510.

\bibitem[{Oono and Suzuki(2019)}]{oono2019graph}
Oono, K.; and Suzuki, T. 2019.
\newblock Graph neural networks exponentially lose expressive power for node
  classification.
\newblock \emph{arXiv preprint arXiv:1905.10947}.

\bibitem[{Rong et~al.(2019)Rong, Huang, Xu, and Huang}]{rong2019dropedge}
Rong, Y.; Huang, W.; Xu, T.; and Huang, J. 2019.
\newblock Dropedge: Towards deep graph convolutional networks on node
  classification.
\newblock \emph{arXiv preprint arXiv:1907.10903}.

\bibitem[{Satorras, Hoogeboom, and Welling(2021)}]{satorras2021n}
Satorras, V.~G.; Hoogeboom, E.; and Welling, M. 2021.
\newblock E (n) equivariant graph neural networks.
\newblock In \emph{International Conference on Machine Learning}, 9323--9332.
  PMLR.

\bibitem[{Veli{\v{c}}kovi{\'c} et~al.(2017)Veli{\v{c}}kovi{\'c}, Cucurull,
  Casanova, Romero, Lio, and Bengio}]{velivckovic2017graph}
Veli{\v{c}}kovi{\'c}, P.; Cucurull, G.; Casanova, A.; Romero, A.; Lio, P.; and
  Bengio, Y. 2017.
\newblock Graph attention networks.
\newblock \emph{arXiv preprint arXiv:1710.10903}.

\bibitem[{Xu et~al.(2018)Xu, Li, Tian, Sonobe, Kawarabayashi, and
  Jegelka}]{xu2018representation}
Xu, K.; Li, C.; Tian, Y.; Sonobe, T.; Kawarabayashi, K.-i.; and Jegelka, S.
  2018.
\newblock Representation learning on graphs with jumping knowledge networks.
\newblock In \emph{International Conference on Machine Learning}, 5453--5462.
  PMLR.

\bibitem[{Yang, Cohen, and Salakhudinov(2016)}]{yang2016revisiting}
Yang, Z.; Cohen, W.; and Salakhudinov, R. 2016.
\newblock Revisiting semi-supervised learning with graph embeddings.
\newblock In \emph{International conference on machine learning}, 40--48. PMLR.

\bibitem[{Zhao and Akoglu(2019)}]{zhao2019pairnorm}
Zhao, L.; and Akoglu, L. 2019.
\newblock Pairnorm: Tackling oversmoothing in gnns.
\newblock \emph{arXiv preprint arXiv:1909.12223}.

\bibitem[{Zhou et~al.(2020)Zhou, Huang, Li, Zha, Chen, and
  Hu}]{zhou2020towards}
Zhou, K.; Huang, X.; Li, Y.; Zha, D.; Chen, R.; and Hu, X. 2020.
\newblock Towards deeper graph neural networks with differentiable group
  normalization.
\newblock \emph{Advances in Neural Information Processing Systems}, 33:
  4917--4928.

\end{thebibliography}
\end{document}